\documentclass[runningheads]{llncs}
\usepackage{graphicx}
\usepackage{comment}
\usepackage{amsmath,amssymb} 
\usepackage{color}
\usepackage{algorithm}

\begin{document}
\pagestyle{headings}
\mainmatter
\def\ECCVSubNumber{3033}  
\title{Relevant-features based Auxiliary Cells for Energy Efficient Detection of Natural Errors} 

\titlerunning{RACs for Energy Efficient Detection of Natural Errors}
%
\author{Sai Aparna Aketi\inst{1} \and
Priyadarshini Panda\inst{1}\and
Kaushik Roy\inst{1}}
\authorrunning{S. A. Aketi et al.}
%
\institute{Purdue University, West Lafayette IN 47906, USA\\
\email{saketi@purdue.edu}}
\maketitle

\begin{abstract}
Deep neural networks have demonstrated state-of-the-art performance on many classification tasks. However, they have no inherent capability to recognize when their predictions are wrong. There have been several efforts in the recent past to detect natural errors but the suggested mechanisms pose additional energy requirements. To address this issue, we propose an ensemble of classifiers at hidden layers to enable energy efficient detection of natural errors. In particular, we append Relevant-features based Auxiliary Cells (RACs) which are class specific binary linear classifiers trained on relevant features. The consensus of RACs is used to detect natural errors. Based on combined confidence of RACs, classification can be terminated early, thereby resulting in energy efficient detection. We demonstrate the effectiveness of our technique on various image classification datasets such as CIFAR-10, CIFAR-100 and Tiny-ImageNet.
\end{abstract}

\section{Introduction}
Machine learning classifiers have achieved high performance on various classification tasks, e.g.,  object detection, speech recognition and image classification. Decisions made by these classifiers can be critical when employed in real-world tasks such as medical diagnosis, self-driving cars, security etc. Hence, identifying incorrect predictions i.e. detecting abnormal inputs is of great importance to safety critical applications. Note that abnormal samples include natural errors, adversarial inputs and out-of-distribution (OOD) examples. Natural errors are samples in the test data which are misclassified by the final classifier of a given network.

Various techniques have been proposed in literature to address the issue of distinguishing abnormal samples. A baseline method for detecting natural errors and OOD examples was proposed in \cite{Hendrycks:2017}. This technique thresholds the Maximal Softmax Response (MSR) in order to detect
natural errors and OOD samples.
A simple unified framework to detect adversarial and OOD samples was proposed in \cite{lee:2018}. They use activations of hidden layers along with a generative classifier to compute Mahalanobis distance \cite{mahalanobis} based confidence score. However, they do not deal with detection of natural errors. 
The authors in \cite{Mandelbaum:2017} use distance based confidence method to detect natural errors.
More recently, the authors in \cite{Bahat:2019} showed that KL-divergence between the outputs of the classifier under image transformations can be used to distinguish correctly classified examples from adversarial and natural errors. 
To enhance natural error detection, they further incorporate Multi Layer Perceptron (MLP) at the final layer which is trained to detect misclassifications. 

Most prior works on the line of error detection do not consider the latency and energy overheads that incur because of the detection mechanism. 
It is known that deeper networks expend higher energy and latency during feed-forward inference. Adding a detector or detection mechanism on top of this will give rise to additional energy requirements. The increase in energy may make these networks less feasible to employ on edge devices. Many recent efforts toward energy efficient deep neural networks (DNNs) have explored \textit{early exit} techniques. Here, the main idea is to bypass (or turn off) computations of latter layers if the network yields high \textit{confidence} prediction at early layers.
Some of these techniques include the adaptive neural networks \cite{dimitrios:2018}, the edge-host partitioned neural network \cite{ko:2018}, the distributed neural network \cite{teera:2017}, the cascading neural network \cite{leroux:2017}, the conditional deep learning classifier \cite{pandap:2016} and the scalable-effort classifier \cite{ramani:2015}.
So far, there has been no unified technique that enables energy efficient inference in DNNs while improving their robustness towards abnormal samples. 

\begin{figure}[ht]
\centering
\includegraphics[width=0.95\textwidth]{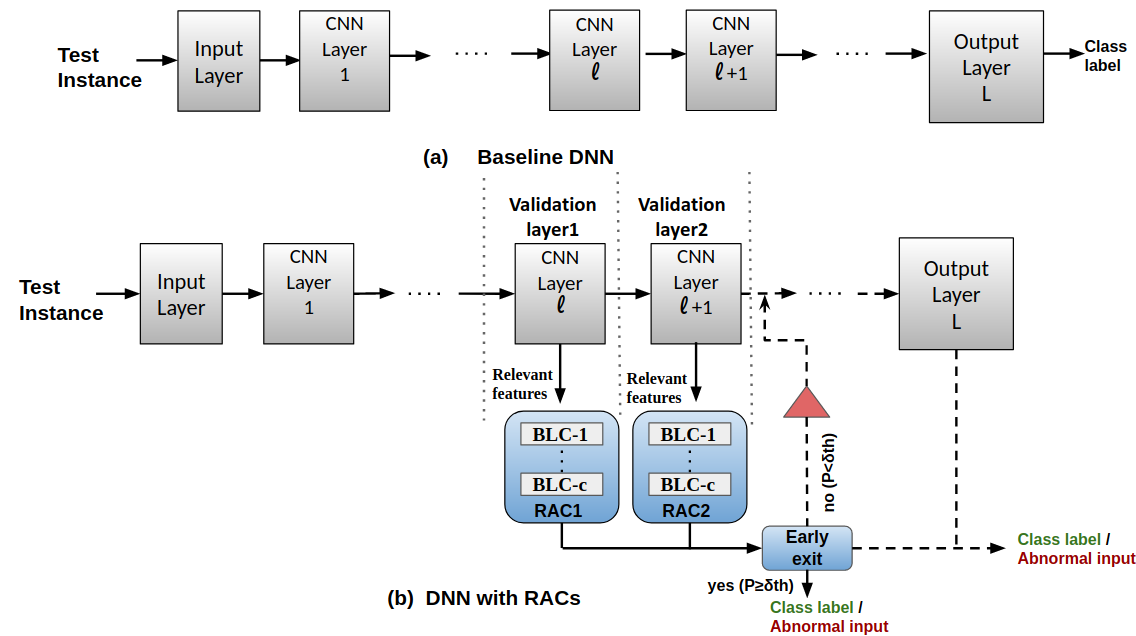}
\caption{(a) Baseline Deep Neural Network (DNN). (b) DNN with Relevant features based Auxiliary Cells (RACs) added at validation layers (selected hidden layers) whose output is monitored to detect early classification.}
\label{fig:main}
\end{figure}

In this work, we target energy efficient detection of natural errors, which can be extended and applied to detecting OOD examples and adversarial data. We propose an ensemble of classifiers at two or more hidden layers of an already trained DNN as shown in Figure.~\ref{fig:main}. 
In particular, we append class-specific binary linear classifiers at few selected hidden layers. These hidden layers are referred to as \textit{validation layers}. The set of all binary linear classifiers at a validation layer constitute a \textit{Relevant feature based Auxiliary Cell (RAC)}. We use the RACs to detect natural errors as well as to perform early classification.
This idea is motivated from the following two observations:
\begin{itemize}
    \item If an input instance can be classified at early layers \cite{pandap:2016} then processing the input further by the latter layers can lead to incorrect classification due to over-fitting. This can be avoided by making early exit which also yields energy efficiency benefits.
    \item We have observed that on an average, the examples which are misclassified  do not have consistent hidden representations compared to correctly classified examples. The additional linear classifiers and their consensus enables identifying this inconsistent behaviour to detect misclassified examples or natural errors.
\end{itemize}

The training and construction of the linear classifiers is instrumental towards the accurate and efficient error detection with our approach. We find that at a given hidden layer, the error detection capability (detecting natural errors) is higher if we use class-specific binary classifiers trained on the corresponding relevant feature maps from the layer. In fact, using a fully connected classifier trained on all feature maps (conventionally used in early exit techniques of \cite{pandap:2016}) does not result in better error detection capability. Training these binary classifiers on relevant features can be considered as encoding prior knowledge on the learned hidden feature maps, thereby, yielding better detection capability. Besides improved error detection, a key advantage of using class wise binary linear classifiers trained on only relevant features is that they incur less overhead in terms of total number of parameters, as compared to a fully connected classifier trained on all feature maps.

We evaluate the efficiency of our methodology on CIFAR-10, CIFAR-100 \cite{alex:2009} and Tiny-ImageNet \cite{tiny} using standard CNN architectures such as VGGNet \cite{szegedy:2015} and ResNet \cite{He:2016}. Our experiments show that for the CIFAR-100 dataset trained on VGG-16 network, RACs can detect 46\% of the misclassified examples along with 12\% reduction in energy compared to the baseline network while 67\% of the examples are correctly classified. This shows that the proposed technique is able to achieve significant reduction in energy along with the decrease in test error.

\subsection{Contribution}
In this paper, we present a novel technique to detect the natural errors in an energy efficient manner which can be extended and applied to adversarial and out-of-distribution samples. We introduce Relevant-features based Auxiliary Cells (RACs) which are class-specific binary linear classifiers trained on relevant features of the corresponding class. We determine the relevant features for each class using an explainable technique called layer-wise relevance propagation \cite{lrp:2015}. The proposed technique constructs an ensemble of classifiers by appending RACs at few selected hidden layers. The inference policy utilizes the consensus of RACs to detect natural errors which improves the robustness of the network. The combined confidence of RACs is used to perform early classification that yields energy efficiency.

\section{Relevance-Score Matrix}
\label{rsv}
DNNs (or convolutional networks) trained for classification tasks compute a set of features at each convolutional layer.
At each layer, few feature maps might highly impact the prediction of a particular class which are considered as relevant features for that class.
For example, a high-level feature can represent $whiskers$ (say) which are relevant to classes like $cat$ and $dog$ but not to classes like $truck$ and $airplane$.
Hence, the feature map computed from this filter is considered as relevant feature for classes $cat$ and $dog$.  
Our approach of adding linear classifiers to trained DNNs follows two steps:
1) First, we heuristically select two or more hidden convolutional layers with maximal information as \textit{validation layers} (refer sec.~\ref{choice of IL}).
2) Then, we determine the class-wise relevant features at the \textit{validation layers} that are eventually used to train the RACs. 

\begin{algorithm}[ht]
\textbf{Input:} Trained DNN, Training data $\{(x_i,y_i)\}_{i=1}^N$:  $x_i\in$ input sample, $y_i\in$ true label\\
\textbf{Parameters:} number of classes: $c$,\hspace{1mm} number of layers: $L$,
feature maps at layer $l$: $\{f_1, f_2, \hdots, f_r\}$,\hspace{1mm} relevance score of node $p$ at
layer $l$ = $R_p^l$ \\

1. Initialize relevance-score matrix for given layer \textit{l}: $M_l=zeros(c,r)$\\
2. \textbf{for} each sample $(x_i,y_i)$ in training data \textbf{do}\\
3.\hspace{0.5cm}Forward propagate the input $x_i$ to obtain the activations of all nodes in the DNN \\
4.\hspace{0.5cm}Compute relevance scores for output layer:
$R_p^L = \delta(p-y_i)\hspace{2mm} \forall p\in \{1,\hdots,c\}$\\
. \hspace{0.5cm} where $\delta(p-y_i) =$ Kronecker delta function \\
5.\hspace{0.5cm}\textbf{for} $k$ in $range(L-1,l,-1)$ \textbf{do}\\
6.\hspace{1.2cm}Back propagation step: 
$R_p^k = \sum_{q}(\alpha \frac{(a_p w_{pq})^{+}}{\sum_p {(a_p w_{pq})^{+}}} -\beta\frac{(a_p w_{pq})^{-}}{\sum_p {(a_p w_{pq})^{-}}})R_q^{k+1}$ \\
.\hspace{1.5cm}$\forall$ $p \in$ nodes of layer $k$; \hspace{1mm} $\alpha - \beta = 1$,
$a_p = activations$, $w_{pq}=weights$, \\
.\hspace{1.5cm}where $()^+$ and $()^-$ denote the positive and negative parts, respectively.\\
7.\hspace{0.5cm} \textbf{end for}\\
8.\hspace{0.5cm} Average relevance scores per feature map:
$R^l = \big\{R^l_{f_j} =\frac{1}{\sum\limits_{p \in f_j}1} \big(\sum\limits_{p \in f_j} R_p^l\big)\big\}_{j=1}^{r}$\\
10.\hspace{0.5cm}Update relevance-score matrix: $M_l(y_i,:) +=R^l $\\
11. \textbf{end for}\\
12. Normalize rows of relevance-score matrix:
$M_l(p,:) = \frac{1}{\sum\limits_{\forall y_i\in p}1} M_l(p,:),$ $\forall p
\in \{1,\hdots,c\}$\\
13. \textbf{return} Relevance-Score Matrix $M_l$\\
\caption{Methodology to Compute Relevance-Score Matrix $M_l$}
\end{algorithm}

To obtain relevant features, we define a relevance-score matrix at each validation layer. It assigns class-wise relevance score to every feature map. The relevance score of a feature map for any given class (say $cat$) indicates its contribution in activating the output node (corresponding to $cat$).
Algorithm 1 shows the pseudo code for computing the relevance-score matrix. The process takes a pre-trained DNN and training data with corresponding labels as inputs and computes relevance-score matrix $M_l$ for a particular layer $l$. 
Each row in $M_l$ indicates the relevance scores of all the features maps at layer $l$ corresponding to a unique class $c$ from the dataset.
In particular, $M_l(i,j)$ indicates the relevance score of feature map $f_j$ at layer $l$ corresponding to class $i$ in the dataset. 

We use \textit{Layer-wise Relevance Propagation (LRP)} proposed in \cite{lrp:2015} to compute the relevance-score matrix.
LRP computes the contribution of every node in the network to the prediction made for an input image.
The relevance scores at output nodes are determined based on true label of an instance. 
For any input sample $(x_i,y_i)$, the output node corresponding to true class, $y_i$, is given a relevance score of $1$ and the remaining nodes get a score of $0$. 
These relevance scores are then back propagated based on $\alpha \beta$-\textit{decomposition rule} \cite{lrp:2016} of LRP with $\alpha=2$ and $\beta=1$. The $\alpha \beta$-\textit{decomposition rule} redistributes the relevance scores from layer $l+1$ to layer $l$ based on the weights and activations of the layers. 
After determining the relevance scores at each node in the network, we compute relevance score of every feature map $f_i$ at layer $l$ by averaging the scores of all nodes corresponding to $f_i$. The relevance vector of a feature map $f_i$ is obtained by taking class-wise average over relevance scores of all training samples and forms the $i^{th}$ column of relevance-score matrix $M_l$.
The computed relevance-score matrix is then used to determine relevant features for each class at validation layers.

\section{\textbf{R}elevant features based \textbf{A}uxiliary \textbf{C}ells (RACs)}
\label{sec:per_class}
In this section, we present our approach to designing DNNs with RACs. 
Fig.~\ref{fig:main} shows the conceptual view of DNNs with RACs. Fig.~\ref{fig:main}(a) consists of the baseline DNN with $L$ layers. We have not shown the pooling layers or the filters for the sake of convenience in representation. Fig.~\ref{fig:main}(b) illustrates our approach wherein the output relevant features from two hidden layer $l, l+1$ which are referred as validation layers are fed to RACs. Note that the two validation layers need not be consequent. 

An RAC consists of $c$ \textit{Binary Linear Classifiers (BLCs)}, where $c$ represents the number of output nodes or the number of classes. Each BLC within an RAC corresponds to a unique class in the dataset and is trained on relevant features corresponding to that class. 
The output of BLC corresponding to a class (say $c1$) in an RAC indicates the probability of a given instance $x_i$ belonging to the class $c1$ and is given $P(y_i=c1|x_i)$.  
We can thus gather that output from an RAC (class label $RAC_{class}$ and associated probability or confidence $RAC_{prob}$) will correspond to the BLC with maximum value as:
\begin{align}
RAC_{class} &= argmax_{i=1,2, \hdots,c} BLC_i \\
RAC_{prob} &= max_{i=1,2,\hdots,c} BLC_i
\end{align}
The probability ($RAC_{prob}$) generated by the RAC is considered as its \textit{confidence score}. 
Besides the RACs, an activation module is added to the network (triangle in Fig.~\ref{fig:main}(b)) similar to that in \cite{pandap:2016}. The activation module uses the consensus of RACs and their confidence scores to decide if an input instance classification can be terminated at the present layer. 

\subsection{Training RACs}
We proceed to train RACs after determining relevance-score matrices (see sec.~\ref{rsv}) at validation layers. Algorithm 2 shows the pseudo code for training RACs. The initial step in this process is to determine the relevant features for each class at the validation layers using relevance-score matrix. For every class $j$, we arrange feature maps in the descending order of their class relevance score and top `$k$' feature maps are marked as relevant features for class $j$.

\begin{algorithm}[ht]
\textbf{Input:} Trained DNN, Training data $\{(x_i,y_i)\}_{i=1}^N$, relevance-score matrix $M_l$\\
\textbf{Parameters:} number of class: $c$\\
1. \textbf{for} each class $j \in 1,\hdots,c$ \hspace{1mm} \textbf{do}\\
2.\hspace{0.5cm}Determine top $k$ relevant features of class $j$ at layer $l$ from $M_l(j,:)$\\
3.\hspace{0.5cm}Obtain relevant features i.e. $x_i^{lj}\hspace{2mm} \forall i \in \{1,\hdots,N\}$ by forward propagating $x_i$ through DNN\\
4.\hspace{0.5cm}Get the binary labels for training data:$\tilde{y}_i = \delta(j-y_i) \hspace{2mm} \forall i \in \{1,\hdots,N\}$ \\
5.\hspace{0.5cm}Initialize a binary linear classifier (BLC-$j$) \\
6.\hspace{0.5cm} Train BLC-$j$ using $\{(x_i^{lj},\tilde{y}_i)\}_{i=1}^N$ as training data and mini-batch stochastic gradient descent algorithm\\
7.\hspace{0.5cm}\textbf{return} BLC-$j$\\
8. \textbf{end for}\\
\caption{Methodology to Train an RAC at Layer $l$}
\end{algorithm}

Once the relevant features for each class are determined, they remain unchanged. The classifier of class $j$ (BLC-$j$) is trained on the corresponding relevant features from the training data. Note, the relevant feature maps which are fed to RACs are obtained after the batch-norm and ReLU operation on selected convolutional layer (validation layer). The BLCs ($BLC-1,...,BLC-c$) in an RAC can be trained in parallel as they are independent of each other. 

\subsection{Early classification and Error detection}
The overall testing methodology for DNNs with RACs is shown in Algorithm 3. We adopt the early exit strategy proposed in \cite{pandap:2016} and modify it to perform efficient classification and natural error detection with RACs. Given a test instance $I_{test}$, the methodology either produces a class label $C_{test}$ or makes \textit{No Decision (ND)}. The output from RACs is monitored to decide if early classification can be made for an input. If the decision made by RACs across the selected validation layers do not agree with each other, then the network outputs \textit{ND} indicating the possibility of misclassification at the final output layer of the DNN. If the RACs across all validation layers predict same class label $c$, then, we use a pre-defined confidence threshold ($\delta_{th}$) to decide on early classification as follows:
\begin{itemize}
    \item If $confidence$ score ($RAC_{prob}$) across all RACs is greater than $\delta_{th}$, we output $c$ as final decision and terminate the inference at the given validation layer without activating any latter layers.
    \item If $confidence$ score ($RAC_{prob}$) in any of the RACs is less than $\delta_{th}$, the input is fed to the latter layers and the final output layer of the DNN is used to make the prediction.
\end{itemize}
In the second case above, all remaining layers from $l+2$ on-wards in Fig. 1(b) will be activated and the output of the final layer ($L$) is used to validate the decision made by RACs. If an input is classified at RACs either as $ND$ or $C_{test}$ (thus, not activating the layers beyond validation layers), then it is considered as an early classification. In Fig. 1, testing is terminated at layer $l+1$ in case of early classification. 

\begin{algorithm}[h]
\textbf{Input:} Test instance $I_{test}$, DNN with RAC-$1$ and RAC-$2$ at validation layers $l$ and $l+1$ respectively \\
\textbf{Output:} Indicates class label ($C_{test}$) or detects abnormal input as No Decision (\textit{ND})\\

1. Obtain the DNN layer features for $I_{test}$ corresponding to layers $l$ and ($l+1$)\\
2. Activate and obtain the output from RAC-$1$ and RAC-$2$\\
3. \textbf{if} RAC-$1_{class}$ $==$ RAC-$2_{class}$ \textbf{do}\\
4.\hspace{0.5cm}\textbf{if} RAC-$1,2_{prob}$ (confidence of each RAC) $> \delta_{th}$ \textbf{do}\\
5.\hspace{1cm}\textit{Terminate} testing at layer ($l+1$)\\
6.\hspace{1cm}Output $C_{test}$ = RAC-$1,2_{class}$ (class label given by RACs)\\
7.\hspace{0.5cm}\textbf{else do}\\
8.\hspace{1cm}Activate remaining layers and obtain prediction ($FC$) from output layer $L$\\
9.\hspace{1cm}\textbf{if} $FC$ $=$ RAC-$1,2_{class}$ \textbf{do} \\ 
.\hspace{1.5cm}Output $C_{test}$ = $FC$\\
10.\hspace{0.9cm}\textbf{if} $FC$ $\neq$ RAC-$1,2_{class}$ \textbf{do} \\
.\hspace{1.5cm}Output \textit{ND}\\
11.\hspace{0.5cm}\textbf{end if}\\
12. \textbf{else do}\\
13.\hspace{0.5cm}\textit{Terminate} testing at layer ($l+1$)\\
14.\hspace{0.5cm}Output \textit{ND}\\
15. \textbf{end if}
  \caption{Methodology to Test the DNN with RACs}
\end{algorithm}

In summary, appending RACs into DNNs enables us to perform early classification with the ability to output a no decision ($ND$) that helps in detecting natural errors (and abnormal inputs). It is evident that early classification will translate to energy efficiency improvements \cite{pandap:2016}. The user defined threshold, $\delta_{th}$, can be adjusted to achieve the best trade-off between efficiency and error detection capability. We believe that the proposed methodology is systematic and can be applied to all image recognition applications.

\section{Experimental Methodology}
\label{sec:exp}
In this section, we describe the experimental setup used to evaluate the performance of DNNs with RACs. 
We demonstrate the effectiveness of our methodology to detect natural errors on state-of-the-art networks, such as VGG \cite{szegedy:2015} and ResNet \cite{He:2016} for image classification tasks on CIFAR \cite{alex:2009} and Tiny-ImageNet \cite{tiny} datasets. We also evaluate our methodology to detect OOD samples and adversarial inputs. For OOD detection, we use LSUN \cite{fisher:2015} , SVHN \cite{yuval:2011} and Tiny-ImageNet datasets as OOD samples for networks trained on CIFAR-10 and CIFAR-100 datasets. In adversary detection, we generate adversarial examples using Carlini and Wagner attack \cite{nicholas:2017} in zero-knowledge and full-knowledge scenarios. Then, we evaluate the performance of our methodology (on CIFAR-10/CIFAR-100 models) against such adversarial inputs.

We measure the following metrics to quantify the performance of our methodology: \textit{\% of good decisions, \% of bad decisions and \% of early decisions}. In case of DNN with RACs, the inputs fall into three different categories : (a) Inputs which are correctly classified (b) Inputs which are classified as $ND$ (c) Inputs which are incorrectly classified.
Note, DNN with RACs output $ND$, when the input can be potentially misclassified at the final output layer ($L$ in Fig. 1 (a)).
The inputs which are either correctly classified or classified as no decision ($ND$) contribute towards \textit{good decisions}. The inputs which are misclassified by the DNN with RACs are considered as \textit{bad decisions}.

We report FNR and TNR to evaluate the error detection capability. In this case, the negatives are the inputs that are misclassified by the baseline DNN and positives are the inputs that are correctly classified by the baseline DNN. TNR is the percentage of misclassified inputs (by baseline DNN) which are classified as $ND$ by DNN with RACs. FNR is the percentage of correctly classified examples (by baseline DNN) which are classified as $ND$ by DNN with RACs.
We do not report the Area Under Receiver Operating Characteristic (AUROC) curve because it requires the detection mechanism to be based on a discriminating threshold (which is usually a distance based measure \cite{Mandelbaum:2017} or maximal softmax response \cite{Hendrycks:2017}). 
Our technique uses consensus to detect the natural errors rather than thresholding a confidence value like other existing techniques \cite{Hendrycks:2017}. Hence, AUROC can not be computed for the proposed technique. 

To measure energy efficiency, we report the normalized number of Floating Point Operations (normalized \#FLOPs). It is the ratio of the average number of FLOPs required by the baseline network to the proposed technique.
We have adopted the PyTorch utility that estimates the number of FLOPs for a given network presented in \cite{flops}. Our goal is to increase TNR and improve energy efficiency while maintaining a low FNR. We observed that the three metrics - TNR, FNR and \#FLOPs are sensitive to hyper-parameters related to RACs and hence, we carried out series of experiments to determine their effect. The details of these experiments are shown in the following section (Sec.~\ref{choice of IL}).

\subsection{Tuning Hyper-parameters}
\label{choice of IL}
Following are three hyper-parameters which affect TNR, FNR and energy efficiency (\#FLOPs): 
\begin{itemize}
    \item The choice of validation layers ($l$, $l+1$)
    \item Number of relevant features ($k$) used at each validation layer
    \item Confidence threshold $\delta_{th}$
\end{itemize}
We use heuristic based methods to tune the above mentioned hyper-parameters using validation dataset.

\begin{figure}[ht]
\centering
\includegraphics[width=1\textwidth]{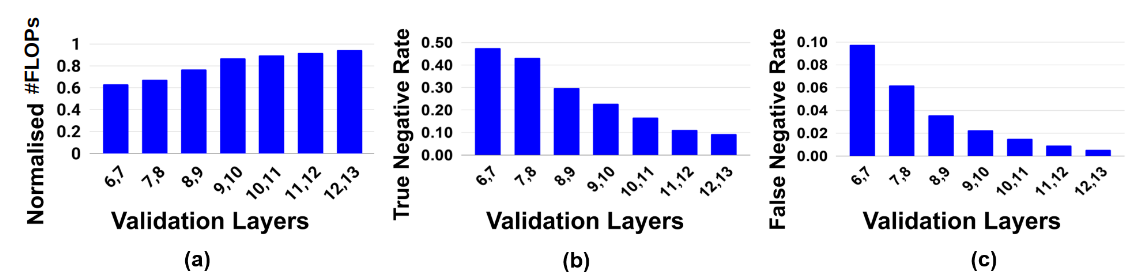}
\caption{(a) Normalised \#FLOPs (with respect to Baseline DNN)- (b) True Negative Rate (TNR) - (c) False Negative Rate (FNR) - shown as the validation layers are shifted towards the final output layer for a VGG-16 network trained on CIFAR-10 dataset.}
\label{fig:bar}
\vspace{-0.5cm}
\end{figure}

\textit{\textbf{Choosing validation layers:}} 
First, lets understand how \textit{validation layers} are chosen and their effect on detection capability. 
The validation layers cannot be initial layers as they do not have the full knowledge of network hierarchy and the feature maps at these layers are not class specific. 
We observed that the hidden layers just before the final output layer (Layer $L$ in Fig. 1) make similar decisions as that of the final output and hence are not useful to detect natural errors. Thus, the hidden layers which are in between (yet, closer to final output) are suitable as validation layers. 
Fig.~\ref{fig:bar} shows the change in FNR, TNR and normalized \#FLOPs with respect to change in the choice of the validation layers for CIFAR-10 dataset trained on VGG-16 network. As validation layers move deeper into the network, both TNR and FNR tend to decrease (Fig.~\ref{fig:bar}(b,c)). 
We select a pair of hidden layers as validation layers which yield low FNR ($5\%-10\%$) with reasonably high TNR ($40\% -50\%$). 
From Fig.~\ref{fig:bar}(b,c), we find that 5\%-10\% FNR range is obtained when we choose \textit{layer 7} and \textit{layer 8} as validation layers for VGG-16 network trained on CIFAR-10 dataset. Similar experiments are done for other network architectures and datasets to choose validation layers that yield an optimum TNR-FNR ratio.

\begin{figure}[ht]
\centering
\includegraphics[width=0.95\textwidth]{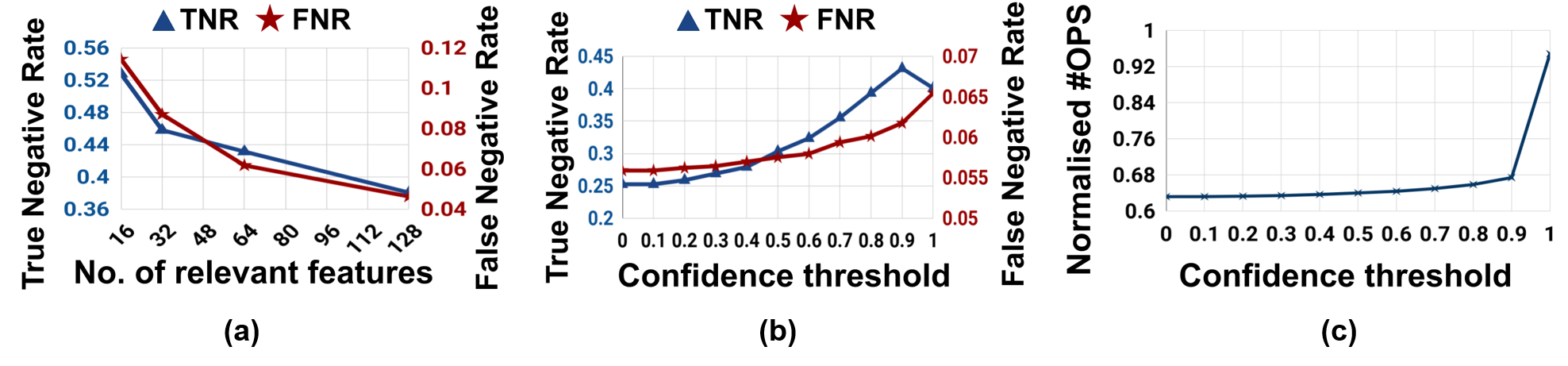}
\caption{(a) TNR and FNR as the no. of relevant features $k$ is increased at RACs (b) TNR and FNR as the confidence threshold $\delta_{th}$ is increased at RACs (c) Normalized \#FLOPs as the confidence threshold $\delta_{th}$ is increased at RACs. All plots shown here are for a VGG-16 network trained on CIFAR-10 with RACs appended at validation layer 7 and 8.}
\label{fig:line}
\vspace{-0.5cm}
\end{figure}

\textit{\textbf{Choosing optimal $k$ relevant features:}} 
Number of relevant features `$k$' is another hyper-parameter which affects FNR/TNR. As we increase the number of relevant features $k$, both FNR and TNR decreases. Fig.~\ref{fig:line}(a) shows the change in FNR and TNR with respect to the change in the number of relevant features $k$ for CIFAR-10 dataset trained on VGG-16 network with validation layers at layer 7 and 8. The optimal value of $k$ depends on the dataset and the network used. 
We increment $k$ by powers of 2, compute the corresponding FNR and TNR and select the optimal $k$ from these experimental observations. Note that \#FLOPs increase as `$k$' increases.
In Fig.~\ref{fig:line}(a), when we increment $k$ from 64 to 128 at the validation layers, the FNR reduces only by 1.6\% dropping from 6.2\% to 4.6\%. In contrast, TNR drops drastically by 5\%. Hence, we choose $k$ as 64 for CIFAR-10 with VGG-16 network to have an optimal TNR-FNR ratio.  

\textit{\textbf{Choosing confidence threshold $\delta_{th}$:}} 
The confidence threshold $\delta_{th}$ is a user defined hyper-parameter which also influences energy efficiency and detection capability.
The activation module discussed in Section.~\ref{sec:per_class} compares the confidence score produced by RACs ($RAC_{prob}$ from Eqn. 2) to $\delta_{th}$ and performs early exit by conditionally activating the latter layers of the network. Thus, we can regulate $\delta_{th}$ to modulate the number of inputs being passed to latter layers. This will in turn impact the overall energy efficiency (or \#FLOPs). Fig.~\ref{fig:line}(c) shows the variation in normalized \#FLOPs (calculated with respect to baseline DNN) for different $\delta_{th}$.  Note that $\delta_{th}$ has no contribution to the decision made when the RACs at different validation layers do not output same class. $\delta_{th}$ also affects the TNR and the FNR. However, the change in FNR with $\delta_{th}$ is negligible. In Fig.~\ref{fig:line}(b)), we observe around 0.01\% change in FNR for 0.1 change in $\delta_{th}$. 

As we increase $\delta_{th}$, TNR increases. Higher $\delta_{th}$ will qualify more inputs to be passed to the latter layers that will eventually get classified by the final output layer. However, beyond a particular $\delta_{th}$, a fraction of inputs which can be correctly classified at early validation stages will be passed to the latter layers and wrongly detected as natural errors because of the increase in confusion. This will adversely affect TNR. In Fig.~\ref{fig:line}(b), we find maximum TNR occurs at $\delta_{th} = 0.9$. It is evident that number of \#FLOPs increase with increasing $\delta_{th}$. In Fig.~\ref{fig:line}(b), we observe that the TNR increases from 39.34\% ($\delta_{th}$=0.8) to 43.15\% ($\delta_{th}$=0.9) while the normalized \#FLOPs increase from 0.66 to 0.67. Going beyond $\delta_{th}$=0.9 degrades the TNR and increases the \#FLOPs by significant amount. Thus, $\delta_{th}$  serves as a knob to trade TNR for efficiency and can be easily adjusted during run-time to get the optimal results.

\section{Results}
This section summarizes results on detection capability and energy efficiency obtained from DNN with RACs.
We train VGG-16 and ResNet-18 for classifying CIFAR-10. For training CIFAR-100 dataset, we use VGG-16 and ResNet-34.
In addition, we also evaluate our approach on ResNet-18 architecture with Tiny-ImageNet dataset.
\begin{table}
\begin{center}
\caption{Baseline network details and the complexity of hidden linear classifiers used for our technique.}
\label{table1}
\begin{tabular}{|c|c|c|c|c|c|}
\hline
\textbf{Dataset} & \textbf{Network} & \textbf{Baseline Error}&\textbf{\# Params} & \textbf{Validation} & \textbf{additional}\\
 & &&  & \textbf{layers} & \textbf{\# params}\\
\hline
CIFAR-10 & VGG-16 & 7.88 & 33.6 M & 7, 8 & 0.08 M\\
\hline
CIFAR-10 & ResNet-18 &5.76  & 11.2 M & 12, 13 & 0.33 M\\
\hline
CIFAR-100 & VGG-16 & 25.62 & 34.0 M & 9, 10 & 0.41 M\\
\hline
CIFAR-100 & ResNet-34& 24.56 & 21.3 M & 31, 32 & 0.82 M\\
\hline
Tiny-Imagenet & ResNet-18 & 43.15  & 11.3 M & 15, 16 & 0.41 M\\
\hline
\end{tabular}
\vspace{-1.5cm}
\end{center}
\end{table}

\subsection{Natural error detection}
Table \ref{table1} indicates the baseline error, the number of parameters in the baseline network, the validation layers used and the additional number of parameters added due to inclusion of RACs.
Table \ref{table:clean} shows the performance of our proposed technique. 
The percentage of early classifications and the energy efficiency is reported in Table.~\ref{table:FLOPs}.
We observe that DNN with RACs can detect  $(43-45)\%$ of the natural errors while maintaining the the percentage of correct decisions at $(86-89)\%$  for CIFAR-10 dataset as shown in Table.~\ref{table:clean}. 

\begin{table}
\begin{center}
\caption{The results show natural error detection performance for DNNs with RACs. We measure the \% of correct classification, \% of $ND$, \% of bad decisions (i.e. the misclassified examples) across different image classification tasks. The FNR, TNR values are also shown. All the values are percentages}
\label{table:clean}
\begin{tabular}{|c|c|c|c|c|c|c|}
   \hline
    {\textbf{Dataset}} & \textbf{Network}& \multicolumn{2}{c|}{\textbf{\% Good decisions}} & \textbf{\% Bad} & \textbf{FNR} & \textbf{TNR} \\
    \cline{3-4}
    & & \textbf{Correct } & \textbf{No} & \textbf{decisions}&\textbf{(\%)} &\textbf{(\%)} \\
    & &\textbf{decisions} & \textbf{decisions}&  \textbf{(Error)}& &\\
\hline
CIFAR-10& VGG-16 & 86.43 & 9.09 & 4.48 & 6.10 & 43.98\\
\hline
CIFAR-10&  ResNet-18& 88.77 & 8.07 & 3.16 & 5.76 & 44.37\\
\hline
CIFAR-100 & VGG-16 & 68.6 & 17.67 & 13.73 & 7.70 & 46.40\\
\hline
CIFAR-100& ResNet-34 &66.72 & 20.98  & 12.3 & 11.19  & 51.04 \\
\hline
Tiny-ImageNet & ResNet-18 & 41.28 & 42.26 & 16.46 & 27.39 & 61.85\\
\hline
\end{tabular}
\end{center}
\end{table}

For CIFAR-100 dataset, we observe slightly higher detection rate for natural errors i.e. $(46-49)\%$ with the percentage of correct decisions ranging from $(67-69)\%$. 
The detection rate is much higher (around $62\%$) for Tiny-ImageNet dataset trained on ResNet-18. However, the percentage of correct decisions drops from $56.85\%$ to $41.28\%$. This can be potentially improved by using deeper networks such as DenseNet \cite{huang2017densely}. Note that the decrease in the percentage of correct decisions is not because of misclassification but is because of false detection and the falsely detected examples fall into the no decision $ND$ category. In other words, certain correctly classified examples now get detected as $ND$. Therefore, even though the percentage of correctly classified examples decrease slightly, we detect $\sim 50\%$ of natural errors compared to the baseline network. 

\begin{table}
\begin{center}
\caption{\#FLOPs efficiency gain of DNN with RACs with respect to baseline DNN and  the percentage (\%) of early classifications for test data, full knowledge adversarial data and Out-of-distribution examples.}
\label{table:FLOPs}
\begin{tabular}{|c|c|c|c|c|c|c|c|}
\hline
\textbf{Dataset} & \textbf{Network}& \multicolumn{3}{c|}{\textbf{Normalized \#FLOPs}} & \multicolumn{3}{c|}{\textbf{\% of early exit}} \\
\cline{3-8}
 &  &\textbf{Test}&\textbf{Adversarial} & \textbf{OOD} &\textbf{Test}&\textbf{Adversarial} & \textbf{OOD} \\
\hline
CIFAR-10 & VGG-16 & 1.48$\times$ & 1.32$\times$ & 1.30$\times$& 88.55\% &66.21\% & 61.97\% \\
\hline
CIFAR-10 & ResNet-18 & 1.22$\times$ & 1.08$\times$ & 1.14$\times$& 74.58\% & 30.26\% & 50.94\% \\
\hline
CIFAR-100 & VGG-16 & 1.12$\times$ & 1.09$\times$ & 1.09$\times$& 87.03\% & 60.53\% & 57.37\% \\
\hline
CIFAR-100 & ResNet-34 & 1.06$\times$ & 1.05$\times$ & 1.06$\times$& 88.45\% & 78.06\% & 84.60\% \\
\hline
\end{tabular}
\vspace{-0.5cm}
\end{center}
\end{table}

\subsection{Robustness towards adversarial and OOD samples}
We also evaluate the detection capability of our approach against adversarial and OOD inputs for CIFAR-10 and CIFAR-100 datasets. 
The adversarial samples are generated using targeted Carlini \& Wagner (CW) attack with $L_2$-norm \cite{nicholas:2017}. We have considered both zero knowledge adversary and full knowledge adversary to evaluate the robustness of DNN with RACs. The zero knowledge adversaries are created such that the attack has (95-100)\% success rate in fooling the baseline DNN.
The mean adversarial distortion (average imposed $L_2$-norm) and adversarial TNR is shown in Table.~\ref{table:adv}. For the zero knowledge evaluation, adversarial TNR indicates the percentage of successful adversaries detected as $ND$. Note that the adversarial examples which can fool the final output of the DNN are considered as successful adversaries in case of zero knowledge attack. 

Full knowledge adversaries are created by including the loss of RACs in the objective function optimized by CW attack. Thus, full knowledge scenario is a stronger attack notion.  We have reported the adversarial detection rate of DNN with RACs for full knowledge adversaries at mean adversarial distortion similar to zero knowledge adversaries. Essentially, the inherent $ND$ output capability of DNN with RACs enables them to detect adversarial inputs in both zero and full knowledge attack scenarios. Note, increasing the mean adversarial distortion or adversarial attack strength in full knowledge case causes a decline in adversarial TNR. Here, training the RACs with both adversarial/clean data as in \cite{advtraining} can lead to better adversarial detection. However, the fact that our methodology requires higher mean distortion to create full knowledge attacks with 100\% success rate establishes its effectiveness for rendering adversarial robustness in DNNs.
\begin{table}
\begin{center}
\caption{Performance of our technique on detecting adversarial data for image classification task. The reported TNR for adversarial input detection is computed at FNR mentioned in Table.~\ref{table:clean}. All the values are percentages.}
\label{table:adv}
\begin{tabular}{|c|c|c|c|c|c|}
\hline
\textbf{Dataset} &\textbf{Network} & \multicolumn{2}{c|}{\textbf{Zero knowledge}} & \multicolumn{2}{c|}{\textbf{Full knowledge}}\\
 \cline{3-6}
  & &\textbf{Adv. TNR}& \textbf{mean $\|.\|_2$} & \textbf{Adv. TNR}& \textbf{mean $\|.\|_2$}\\
\hline
CIFAR-10 & VGG-16 & 38.42 & 1.32  & 57.35 & 1.33 \\
\hline
CIFAR-10 & ResNet-18 & 44.76  & 1.38 & 9.10 & 1.35 \\
\hline
CIFAR-100 & VGG-16 &37.39 &1.01 & 28.39 & 1.01 \\
\hline
CIFAR-100 & ResNet-34 & 43.49 & 0.79 & 13.95 & 0.73 \\
\hline
\end{tabular}
\vspace{-1cm}
\end{center}
\end{table}

\begin{table}
\begin{center}
\caption{Performance of our technique on detecting OOD samples for image classification task. The reported TNR for OOD samples detection is computed at FNR mentioned in Table.~\ref{table:clean}. All the values are percentages.}
\label{table:ood}
\begin{tabular}{|c|c|c|c|c|}
\hline
\textbf{Dataset} & \textbf{Network} &\textbf{Tiny-ImageNet}&\textbf{LSUN} & \textbf{SVHN} \\
\hline
CIFAR-10 & VGG-16 & 44.25 & 48.10 & 63.96 \\
\hline
CIFAR-10 & ResNet-18 & 60.55 & 68.46 & 70.13\\
\hline
CIFAR-100 & VGG-16 & 43.03  & 38.28  & 38.02\\
\hline
CIFAR-100 & ResNet-34 & 58.55 & 60.30 & 58.56\\
\hline
\end{tabular}
\vspace{-0.5cm}
\end{center}
\end{table}

The detection capability of RACs in case of OOD examples is shown in Table.~\ref{table:ood}. We summarize the percentage of inputs classified early at the validation layers for different datasets and networks in Table.~\ref{table:FLOPs}. 
We see that across all kinds of abnormal inputs, $>50\%$ of the data are classified early in the validation layers by the RACs. This will eventually translate to energy efficiency (see Fig. 5, Table~\ref{table:FLOPs}). We find that the efficiency gain in terms of \#FLOPs is $1.06-1.30 \times$ for OOD detection and $1.05-1.32 \times$ for adversary detection (see details in Table~\ref{table:FLOPs}). The proposed DNN with RAC technique, thus, detects natural errors, adversarial and OOD examples with good TNR-FNR ratio while being energy efficient.

\subsection{Comparison with other existing techniques}
We compare our technique with the existing natural error detection technique proposed in \cite{Hendrycks:2017}. The authors in \cite{Hendrycks:2017} use the Maximal Softmax Response (MSR) in order to detect natural errors. Table.~\ref{table:comp} shows the average reduction in the number of FLOPs and the detection capability reported as TNR for both DNN with RACs and MSR techniques as compared to the baseline network.
We observe that the detection capability of MSR is $(10-20)\%$ higher than DNN with RACs at similar FNR. However, DNN with RACs has better adversarial error detection capability ($(1-10)\%$ higher) and  yields energy efficiency gain of $(5-35)\%$ than MSR.
Thus, the proposed technique has a trade off between the error detection capability and energy efficiency. 
\begin{table}
\begin{center}
\caption{Comparison of the proposed technique (DNN with RACs) with the MSR technique proposed in \cite{Hendrycks:2017}. The TNR for both the techniques is reported at same FNR values (which are reported in Table.~\ref{table:clean}).}
\label{table:comp}
\begin{tabular}{|c|c|c|c|c|c|c|c|}
\hline
\textbf{Dataset} & \textbf{Network}& \multicolumn{2}{c|}{\textbf{Avg. reduction}} & \multicolumn{2}{c|}{\textbf{Natural error TNR}} & \multicolumn{2}{c|}{\textbf{Adversarial TNR}}\\
 & & \multicolumn{2}{c|}{\textbf{in \#FLOPs}} & \multicolumn{2}{c|}{} & \multicolumn{2}{c|}{\textbf{Zero-knowledge}}\\
\cline{3-8}
 &  &\textbf{RACs} & \textbf{MSR} &\textbf{RACs}&\textbf{MSR}&\textbf{RACs}&\textbf{MSR} \\
\hline
CIFAR-10 & VGG-16 & 33\% & 0\% &  44\% & 57\%& 38\%& 29\%\\
\hline
CIFAR-10 & ResNet-18 & 18\% & 0\% & 44\% & 66\%& 45\%& 41\%\\
\hline
CIFAR-100 & VGG-16 & 12\% & 0\% & 46\%& 51\%& 37\%& 33\%\\
\hline
CIFAR-100 & ResNet-34 & 6\% & 0\% & 51\% & 63\%& 44\%& 43\%\\
\hline
\end{tabular}
\vspace{-0.5cm}
\end{center}
\end{table}

The authors in \cite{Bahat:2019}, \cite{Mandelbaum:2017} also focus on detecting natural errors and have reported better detection rate than MSR. But, both these techniques impose enormous energy requirements. The most recent work by Bahat et al. \cite{Bahat:2019} use KL-divergence between the outputs of the classifier under image transformations. For each image, they create `m'$(\geq 2)$ transformed images,  pass it through the network and the resulting top N logits from original and transformed images are used to detect natural errors. This requires to compute $m\times$ more FLOPs than MSR and nearly $1.5m\times$ more FLOPs than DNNs with RACs.
Hence, DNN with RAC approach yields competitive detection capability as compared to MSR with significant compute reduction.

\section{Conclusion}
\label{conclusion}
Deep neural networks are crucial for many classification tasks and require robust and energy efficient implementations for critical applications. In this work, we devise a novel post-hoc technique for energy efficient detection of natural errors. In essence, our main idea is to append class-specific binary linear classifiers at few selected hidden layers referred to as Relevant features based Auxiliary Cells (RACs), which enables energy efficient detection of natural errors. With explainable techniques such as Layer-wise Relevance Propagation (LRP), we determine relevant hidden features corresponding to a particular class which are fed to the RACs. The consensus among RACs and the final output layer is used to detect natural errors. The confidence of RACs is utilized to decide on early classification which yields compute \#FLOPs reduction. We also evaluate the robustness of DNN with RACs towards adversarial inputs and out-of-distribution samples. Beyond the immediate application to increase robustness towards natural errors and reduce energy requirement, the success of our framework suggests further study of energy efficient error detection mechanisms using hidden representations.

\clearpage
\bibliographystyle{splncs04}
\bibliography{egbib}
\end{document}